# A Robust BERT-Based Deep Learning Model for Automated Cancer Type Extraction from Unstructured Pathology Reports.

Minh Tran [*1], Jeffery C. Chan [1], Min Li Huang [1,3,5], Maya Kansara [1,5], John P. Grady [5], Christine E. Napier [5], Subotheni Thavaneswaran [1,2,3,6], Mandy L. Ballinger [5], David M. Thomas [5], Frank P. Lin [1,2,3]

[1] *Faculty of Medicine and Health, University of New South Wales, NSW, Australia*
[2] *Garvan Institute of Medical Research, Darlinghurst, NSW, Australia*
[3] *NHMRC Clinical Trials Centre, University of Sydney, UNSW, Australia*
[4] *SydPath, St Vincent's Hospital, Sydney, NSW, Australia*
[5] *Centre for Molecular Oncology, University of New South Wales, NSW, Australia*
[6] *The Kinghorn Cancer Centre, St Vincent's Hospital, Darlinghurst, NSW, Australia*

**Abstract.**

The accurate extraction of clinical information from electronic medical records is particularly critical to clinical research but require much trained expertise and manual labor. In this study we developed a robust system for automated extraction of the specific cancer types for the purpose of supporting precision oncology research. from pathology reports using a fine-tuned RoBERTa model. This model significantly outperformed the baseline model and a Large Language Model, Mistral 7B, achieving F1_Bertscore 0.98 and overall exact match of 80.61%. This fine-tuning approach demonstrates the potential for scalability that can integrate seamlessly into the molecular tumour board process. Fine-tuning domain-specific models for precision tasks in oncology, may pave the way for more efficient and accurate clinical information extraction.

## 1. Introduction

Diagnostic information contained in the anatomical pathology reports, such as cancer type, grade, and morphology, are critical in optimizing the care of cancer patients [1]. In clinical research, information extracted from histopathology reports is vital, but often requires considerable time and resources to perform data extraction with manual extraction prone to human errors, variations in the skills of the trained extractors. While digitalized pathology reports offer opportunities for automating the extraction process for supporting research, the scale-up efforts are limited by the presentation of information in an unstructured, free-text format, compounded by paper-based reports that are often only digitalized into raster images [2]. The variability in clinical terminology further complicates the extraction process, making it difficult to enhance the quality of extraction at scale.

Machine learning approaches have been developed to extract cancer-related information from pathology reports. Differing from hand-crafted rules, recurrent neural networks (RNNs) [3], Long Short-term memory (LSTM), pre-trained bidirectional encoder representations transformers model (BERT), and more recently, Large language models (LLMs), such as GPT and LLaMA [12], have shown great potential in the field of medical information retrieval in processing unstructured text data.

---
[1] Corresponding Author: Minh Tran, minh_tran@unsw.edu.au
* Co-first authors

Notably, BERT-based models have demonstrated exceptional performance in named entity recognition (NER) tasks to extract clinical information from unstructured medical texts [4, 5]. Various BERT-based models such as BioBert, [6], ClinicalBert [7], and SciBert [8] have been developed to cater for domain specific needs. RoBERTa, stands for Robustly Optimized BERT pretraining approach, sharing the same robust architecture of BERT [9, 10]. These models can process and understand vast amounts of unstructured medical data. They can identify and categorize medical entities through prompt queries, allowing LLMs to adapt to new tasks with minimal training data, making them particularly useful in scenarios where annotated datasets are scarce.

In this study, we developed domain-specific language models and fine tuned a RoBERTa based model to address the challenges of extracting of cancer types from pathology reports. Deep learning models pretrained on general corpora have shown potential, but they encounter significant challenges when applied to extract cancer information from pathology reports. These limitations stem from vocabulary mismatches, subtle nuances in clinical language, and heterogeneous formats of pathology reports. Our approach addresses these gaps by tailoring the model to unique requirements of cancer type extraction, demonstrating the necessity of domain specific fine turning for achieving high accuracy and robustness in this field.

## 2. Methods

*Source data*

To develop an natural language processing (NLP) pipeline to assist with automated extracting cancer diagnoses from unstructured pathology reports, we examined data from an Australia-wide precision oncology program, the Australian Molecular Screening and Therapeutic (MoST) protocol [11], a national precision oncology cohort that uses genomic profiling to identify potential therapeutic opportunities for patients (Australian Clinical Trials Registry ACTRN12616000908437). Text corpora were generated from anatomical pathology reports as part of the referral process. Each diagnosis was classified by International Classification of Diseases for Oncology (ICD-O-3) triplets (comprising cancer type, topography, and morphology). A medical oncologist manually mapped the ICD-O-3 triples to a standardized vocabulary ontology, designed to assist symbolic reference of knowledge-based inference for therapy recommendations [12, 13] to generate a gold-standard dataset. This ontology incorporated two hierarchical classifications: a broad cancer type designation (e.g. colorectal cancer) and a normalized nomenclature that describes the specific histologic subtypes (e.g. colon adenocarcinoma). The data cut-off for the analysis was 20 July 2022.

*Data preprocessing*

De-identified pathology reports were extracted from the institutional database repository, which comprises scanned or facsimile images in Portable Document Format (PDF) files. An optical character recognition (OCR) pipeline was first applied, based on the DocTR backbone, using ResNet-50 for text detection and VGG-15 for word recognition [14] This was followed by spell correction and partitioning into text chunks, based on the relative positioning of the text blocks, with clinically irrelevant information (i.e. headers and footnotes) removed.

These labels were extracted from the paragraphs where the exact matches were found (3,489 paragraphs). Given the variation between standard diagnosis and actual clinician writing, either broad or specific cancer types were not always found in exact match in every report (3091 paragraph). To account for this variation in clinical writing styles, we manually annotated the 161 paragraphs where the official labels were written differently from actual pathologist diagnosis (e.g., "met prostatic adenocarcinoma" instead of "metastatic prostate cancer"). We hypothesized that a sufficiently robust model, with substantial training would be capable of accurately retrieving the cancer types from the input text. Ultimately, our dataset comprised of 3634 records, which were split into training, validation, and test sets in proportions of 70%, 10% and 20% respectively.

*Language Models and Fine-tuning strategy*

We selected RoBERTa as the main architecture for its compelling performance in capturing medical nuances and contextual understanding in domain adaptation. Its underlying improvements stem from training with larger mini-batches and optimized learning rates, making it particularly well-suited for applications requiring deeper contextual understanding. Compared with the baseline model, a fine-tuned RoBERTa model on a question-answering task to find answers for two questions: *"Which cancer is mentioned?"* and *"What is the specific cancer type?"*. The performance of RoBERTa and a fine-tuned RoBERTa were benchmarked against an open-source autoregressive LLM (Mistral) model [15], applying text generation for diagnosis extraction leveraging the scalable linguistic applications and advanced information retrieval. Due to computational and data constraints, the 7-billion-parameter Mistral model was used with a standardized prompt template comprising instructions, context (dataset paragraphs), and a request to extract cancer types and subtypes. To preserve the privacy of the patient data, we cannot use the commercialized language models such as GPT4 or Claude. The same context-question pairs were used for consistency across models during fine-tuning and benchmarking.

*Informatics analyses and performance evaluation*

We compared the performance of the three models using standard exact match, macro-averaged F1 score and F1-BERTScore. Exact match and macro F1 score assess the percentage of predictions that either perfectly match or overlap with any of the correct answers. While exact match and standard macro-F1 score account for partial correctness to an extent, both metrics cannot identify semantic closeness or paraphrased responses that do not involve work overlap. Given that cancer diagnoses can appear in various forms within the context, we considered semantic closeness a critical evaluation criterion. To address this, we employed BERTScore, which uses pre-trained contextual embeddings to measure the similarity between model predictions and ground truth [16]. BERTScore has been shown to correlate strongly with human judgment in assessing paraphrases and provides F1-BERT measures based on the embedded prediction-answer pairs [16].

## 3. Results

*Patients and clinical characteristics*

A total of 5,750 free-text pathology reports were analyzed from 3,322 of 4,887 patients (67.9%) screened through the MoST program, all of whom had suitable anatomical pathology reports available for analysis. This mapped 43 broad cancer types and 288 unique subtypes. The most common cancer types were sarcoma (607 cases, 18.3%), colorectal cancer (350 cases, 10.5%), pancreatic cancer (306 cases, 9.2%), and gliomas (251 cases, 7.6%). Additionally, 1,537 patients had multiple pathology reports with more than one primary diagnosis.

*Performance of BERT models*

The evaluation metrics comparing the model's prediction with the gold standard are shown in Table 2. The baseline RoBERTa model achieved an exact match accuracy of 20.4%, a macro-averaged F1 score of 0.39, and an F1-BERT score of 0.85. While its semantic understanding of the baseline model was reasonably strong, its exact match and macro-F1 scores were considerably lower, indicating limitations in precisely identifying and overlapping with the actual answer. Our fine-tuned RoBERTa model demonstrated significant improvements across all metrics compared to the general RoBERTa model, achieving an exact match accuracy of 80.61%, and an F1-BERT score of 0.98. In contrast, the Mistral-7B model yielded the least accuracy performance at 14.7% matched answer and a macro-averaged F1 score of 0.39. Although both the baseline RoBERTa and Mistral-7B models demonstrated relatively high semantic similarity scores (F1-BERT), they struggled with exact matches and macro F1scores, indicating that immediate use of these models is insufficient for this specialized task without additional domain-specific adaptation (Table 2).

Error analysis revealed that the general-trained RoBERTa frequently failed to extract specific cancer subtypes accurately. Mistral-7B, while somewhat better at extracting specific subtypes, often provided the same answer for different questions. This indicates that baseline RoBERTa and Mistral-7B struggled to discern the differences in oncology terminology. This emphasizes the necessity of domain adaptation and fine-tuning to enhance the model's understanding of specialized vocabulary and nuanced distinctions critical for oncology application.

Table 2. The comparison between three different models in cancer subtype extraction by general RoBERTa, fine-tuned RoBERTa and Mistral-7B.

| Language Models | Exact match | Macro-Averaged F1 | F1_BERT |
| --- | --- | --- | --- |
| RoBERTa | 20.4% | 0.39 | 0.85 |
| Fine-tuned Roberta | **80.61%** | **0.85** | **0.98** |
| Mistral-7B | 14.7% | 0.39 | 0.77 |

The fine-tuned model is available on Hugging Face at: https://huggingface.co/DrM/Cancer_subtype_extraction_from_context and the full pipeline for extraction is available at: https://huggingface.co/spaces/DrM/cancer_subtype_extraction

## 4. Discussion

We developed a system that identifies and efficiently categorizes diverse cancer types for supporting precision oncology research. While deep learning approaches require large, well-annotated datasets in specialty domains, which presents challenges in clinical application, our study highlights several notable strengths. The development and validation of our pipeline within a national precision oncology framework provided access to an extensive and phenotypically diverse patient cohort. Notably, the MoST program has a high representation of rare malignancies, such as sarcoma. A key innovation of our approach was the integration of OCR technology to process image-based pathology reports of varying quality and format, while achieving robust F1 and F1-BERT scores in the automated extraction of cancer types. Furthermore, our BERT-based pipeline mapped unstructured clinical narratives to a standardized ontology at the most specific hierarchical level. This automated approach represents a significant advancement toward scalable processing and workflow improvement, with the potential to assist in decision support systems including molecular tumor board workflows.

**Acknowledgements**

The MoST Program received funding support from the Australian Federal Government and the Office for Health and Medical Research (NSW, Australia). MT and FL are supported by fellowships from *Maridulu Budyari Gumal* (SPHERE) Cancer Clinical Academic Groups (Cancer Institute NSW Research Capacity Building Grant

2021/CBG003). FL and JC are also supported by the Australian Medical Research Future Fund 2021 Rapid Applied Research Translation Grant (RARUR00125).